\pdfoutput=1

\documentclass[11pt]{article}

\usepackage[]{ACL2023}

\usepackage{times}
\usepackage{latexsym}
\usepackage{amsmath}

\usepackage[T1]{fontenc}

\usepackage[utf8]{inputenc}
\usepackage{microtype}
\usepackage{inconsolata}
\usepackage{graphicx}

\usepackage{amssymb}
\usepackage{latexsym}
\usepackage{array}
\usepackage{subcaption}
\usepackage{easytable}
\usepackage{booktabs}

\title{Table Transformers for Imputing Textual
Attributes}

\author{Ting-Ruen Wei \\
  Santa Clara University \\
\And
  Yuan Wang \\
  Santa Clara University \\
\And
Yoshitaka Inoue \\
  DOCOMO Innovations \\
\AND
  Hsin-Tai Wu \\
  DOCOMO Innovations \\
\And
Yi Fang \\
  Santa Clara University \\
  }

\begin{document}
\maketitle
\begin{abstract}
Missing data in tabular dataset is a common issue as the performance of downstream tasks usually depends on the completeness of the training dataset. Previous missing data imputation methods focus on numeric and categorical columns, but we propose a novel end-to-end approach called Table Transformers for Imputing Textual Attributes (TTITA) based on the transformer to impute unstructured textual columns using other columns in the table. We conduct extensive experiments on three datasets, and our approach shows competitive performance outperforming baseline models such as recurrent neural networks and Llama2. The performance improvement is more significant when the target sequence has a longer length. Additionally, we incorporate multi-task learning to simultaneously impute for heterogeneous columns, boosting the performance for text imputation. We also qualitatively compare with ChatGPT for realistic applications.
\end{abstract}

\section{Introduction}
Data quality is crucial for models to make valuable inferences \citep{Schelter2021}. With automated data collection pipelines such as implicit feedback, the growing amount of data can facilitate training, but the quality of data remains a topic of interest. Specifically, datasets in the real world have missing values, and simply dropping that column from the equation is sub-optimal since many variables can have a large correlation with the target column. Moreover, dropping the entire row with missing values dismisses the other completed columns, and doing so can under-represent such populations, which harms the value of the model for those groups. Missing data is categorized into three types: missing completely at random (MCAR), missing at random (MAR), and missing not at random (MNAR). MCAR defines the situation where the absence of data is random and independent from the other variables. MAR describes when the missingness remains random but can be accounted for by other variables, and other situations fall into the MNAR category where missingness is not random. Modern practices dealing with missing data include replacing missing values with the mean, the mode \citep{pratama2016review}, some other constants, or replacing them with the predictions from another machine learning model \citep{alabadla2022systematic} trained on existing data. These are suitable for numeric and categorical missing columns but not for unstructured textual columns which can be an important piece of information in many datasets such as user reviews. Many related works have attempted to solve the missing data problem \citep{mostafa2019imputing,khan2020sice}, but there has been little research on imputing textual columns. To fill that gap, we propose a transformer-based approach to impute unstructured text. The model takes in other heterogeneous columns in the table, including numeric, categorical, and textual inputs and encodes them into a context vector. This vector provides background information on the target column, and the entire pipeline is learned in an end-to-end fashion. Thus, our contribution can be summarized as the following:
\begin{itemize}
    \item We identified the missing data imputation problem for unstructured textual columns in tabular data. To the best of our knowledge, no prior work has addressed the task of table imputation involving unstructured textual attributes.
    \item We proposed a transformer-based model called Table Transformers for Imputing Textual Attributes (TTITA). The model is customized to encode observed heterogeneous tabular attributes as input and decode missing textual attributes as output.
    \item We further conducted experiments on real-world datasets to demonstrate the effectiveness of our model.
    \item We made TTITA open-source as a PIP package for custom applications. Code is released\footnote{https://github.com/tingruew/TTITA-Text-Imputation}.
\end{itemize}

\begin{figure*}[t]
  \centering
  \includegraphics[width=\linewidth]{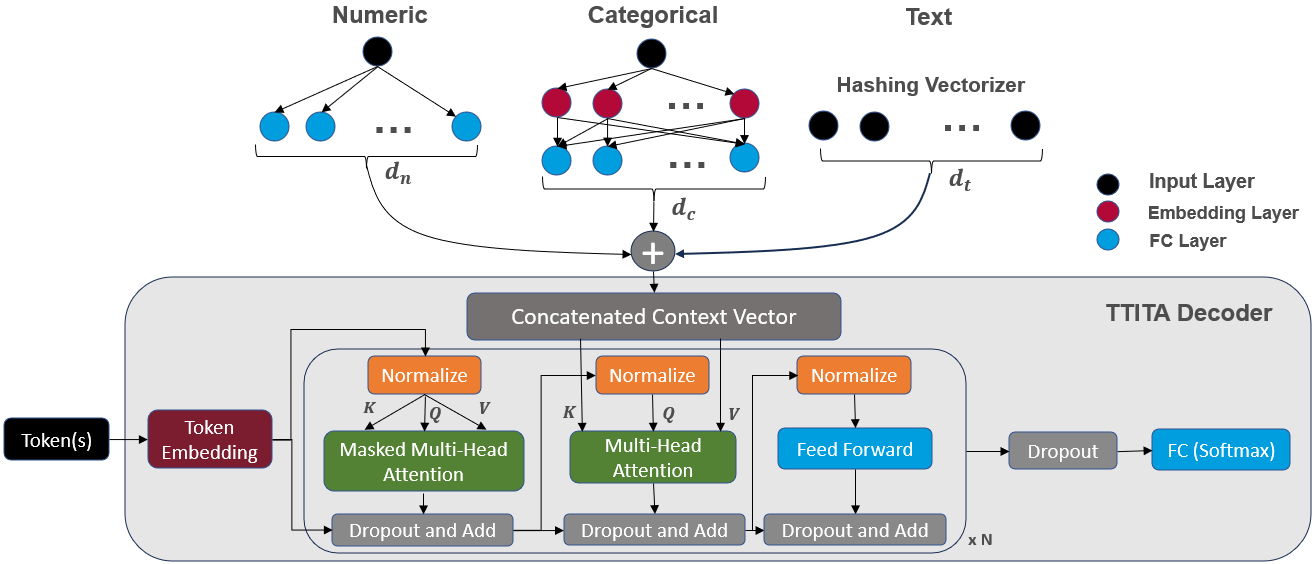}
  \caption{The architecture of the proposed TTITA Model. Input columns are encoded as the following: numeric columns connect to an FC layer of dimension $d_n$, categorical columns connect to an embedding layer and then an FC layer of size $d_c$, and textual columns are featurized by a hashing vectorizer of dimension $d_t$. The latent representations are concatenated and fed into TTITA's decoder for cross-attention, which generates the imputed sequence in an auto-regressive manner.}
  \label{transformer}
\end{figure*}

\section{Related Work}
The data imputation problem has been widely studied, from traditional machine learning methods to large language models (LLMs), but little research has been conducted on imputing unstructured textual data.
DataWig \citep{biessmann2018deep,biessmann2019datawig}, a working software package that motivated our work, can take numeric, categorical, and textual input columns to impute numeric or categorical columns, using embedding and fully-connected layers. \citet{batra2022pragmatic} used an ensemble strategy to impute numeric columns, while \citet{li2021variational}, \citet{mattei2019miwae}, and \citet{gondara2018mida} used variants of autoencoders for imputing time series, image, numeric and categorical data. \citet{jager2021benchmark} and \citet{paterakis2024we} compared various imputation methods for categorical and numeric columns, and \citet{yoon2018gain} and \citet{wu2023differentiable} utilized generative adversarial networks to impute numeric columns. With the recent diffusion model, DiffImpute \citep{wen2024diffimpute}, MTabGen \citep{villaizan2024diffusion}, TabDDPM \citep{kotelnikov2023tabddpm}, TabCSDI \citep{zheng2022diffusion}, and \citet{jolicoeur2024generating} worked on the imputation of numeric and categorical columns. \citet{mei2021capturing} leveraged pre-trained language models to impute numeric and categorical columns, while \citet{narayan2022foundation} showed that language foundation models are able to conduct data cleaning and integration tasks and RetClean \citep{ahmad2023retclean} leveraged ChatGPT to impute missing categorical data with a user-provided data lake. RPT \citep{tang2021rpt} fills in missing data with transformers by concatenating all columns for a data sample into a sequence and reconstructing it. However, unstructured textual columns significantly increase the length of the input and output sequences, which poses a challenge in performance and efficiency to the denoising approach. Under the scope of these related work, our proposed approach, TTITA, tackles the text imputation problem that is highly underdeveloped and specializes in unstructured text imputation for the first time. By using a feature extractor whose output dimension is independent from the input text length, the sequence length of the input textual columns does not hinder the generation of text.

\section{Proposed Method}
Due to the many advantages and the public acceptance of transformer models \citep{vaswani2017attention}, we propose a transformer-based approach using other heterogeneous tabular columns as input to impute textual columns. Input columns are encoded into a context vector that is subsequently fed into the transformer decoder network for cross-attention. Being the query in cross-attention, previous tokens in the target column highlight the relevant parts of the context vector, assisting the token prediction task in the transformer decoder. TTITA has an encoder-decoder architecture where the encoder encodes the input columns and the decoder outputs the text sequence for imputation, facilitated by the cross-attention. Training on domain data exempts from the post-processing that LLMs, including ChatGPT, require for the final result, and the use of featurizers and embeddings enables flexible adjustment on the model size, making this framework adoptable for wide usage.

\subsection{Input Data Encoding}
For each type of input column, we have a different way of encoding in the neural network, as shown in Figure \ref{transformer}. Following DataWig \citep{biessmann2018deep, biessmann2019datawig}, numeric input columns are encoded by a fully-connected (FC) layer of $d_n$ neurons, categorical input columns are connected to an embedding layer and sequentially another FC layer of $d_c$ neurons, and textual input columns are featurized by the hashing vectorizer from the scikit-learn package in a dimension of $d_t$. This corresponds to the size of the input layer for textual columns and does not require any pre-processing of textual data. At the character level, the hashing vectorizer scans for unigrams through 5-grams. We acknowledge that, compared to data imputation, the primary focus lies in the downstream tasks, and the time cost in imputation plays a crucial role in selecting the appropriate imputation method, so we choose not to introduce additional parameters here given the presence of a transformer decoder in the model.

The latent features encoded for all input columns are concatenated into a flat context vector $V_c$ that is connected to the decoder network for end-to-end learning. The context vector $V_c$ is defined as:

\small
\begin{gather}
 \label{context vector}
   V_c=(N_0,...,N_{d_n-1},C_0,...,C_{d_c-1},T_0,...,T_{d_t-1})
 \end{gather}

\normalsize

 \noindent where $N_m$, $C_m$, and $T_m$ are the $m^{th}$ latent feature encoded for numeric, categorical, and textual input columns, respectively. 

 The framework does not require training data to be a complete table. In the training set, missing values in numeric, categorical, and textual columns are replaced by the mean, the ``missing'' category, and an empty string, respectively. As opposed to removing rows with missing values from the training set, this promotes the generality of the framework at the potential expense of performance.

\subsection{Decoder Model}
The imputation of textual attributes is achieved by the transformer decoder part of TTITA, as shown in Figure \ref{transformer}. After encoding the input data into the context vector, TTITA uses it as the key and value vectors in the attention mechanism. The dimension of the token embedding layer is identical to the size of the context vector. Instead of absolute positional encoding, we leverage rotary positional encoding \citep{su2024roformer} to enhance the positional information of each token. We utilize RMSNorm \citep{zhang2019root} for normalization and adopt such layers before the attention operations for better performance. The first attention mechanism draws connections to all parts of the input sequence, countering the issue of long-range dependency that recurrent neural networks have. The second multi-head attention block incorporates the context vector for cross-attention, with attention $A$ calculated as:

 \begin{equation}
 \label{attention}
   A=softmax(\frac{QV_c^T}{\sqrt{d_{V_c}}})V_c
 \end{equation}

\noindent where $Q$, $V_c$, and $d_{V_c}$, are the query, context vector, and the size of the context vector. The Dropout and Add block applies a dropout layer and adds up the two inputs. Instead of the rectified linear unit, we select sigmoid linear unit \citep{hendrycks2016gaussian} as the gated activation function in the Feed Forward block which consists of two FC layers. The entire decoder layer from the first normalization to the last dropout layer and addition is repeated $N$ times before the final dropout layer and token classification.

Another advantage of transformers is the parallel computation that enables higher efficiency. During training, masking prevents foreseeing the subsequent tokens at each position and during inference, the predicted token is concatenated with the input token(s) in an iterative fashion until the $[end]$ tag is predicted or when the maximum length is reached, to obtain the imputed sequence.

\begin{figure}[t]
  \centering
  \includegraphics[width=\linewidth]{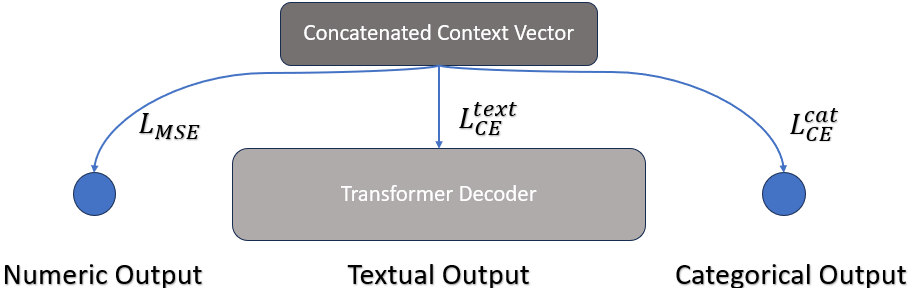}
  \caption{Structure of multi-task learning. We adapt TTITA to a multi-task learning framework that trains to simultaneously impute multiple columns of numeric, textual, or categorical types.}
  \label{mtl}
\end{figure}

\subsection{Multi-task Learning}
To further enhance the imputation of textual attributes, we leverage multi-task learning \citep{caruana1997multitask} to simultaneously impute multiple columns in a table. Multi-task learning has shown great success in computer vision \citep{yim2015rotating} and natural language processing \citep{luong2015multi} by learning the representation suitable for multiple related tasks. Here, we jointly train to impute multiple columns, making an effort to enhance the context vector that provides the context for our primary goal of textual imputation. This not only enables the imputation of multiple columns at once, but also refines the context vector, providing a more comprehensive context for text imputation. The framework considers output columns to be numeric, categorical, or textual, but we focus on the performance of text imputation as numeric and categorical imputation has been widely studied. Figure \ref{mtl} illustrates the multi-task learning framework; the context vector (from the one in Figure \ref{transformer}) connects to different branches, one for each target column. For loss functions, we use mean squared error $L_{MSE}$ defined as:

\begin{equation}
 \label{mse}
   L_{MSE}(y_i, \hat{y}_i)=\sum^{N}_{i=1}\frac{(y_i-\hat{y}_i)^2}{N}
 \end{equation}

\noindent where $N$ is the total number of instances, $y_i$ is the ground truth for the $i^{th}$ instance, and $\hat{y}_i$ is the prediction of the $i^{th}$ instance for numerical target columns. For categorical columns, the loss function is the categorical cross-entropy $L_{CE}^{cat}$ defined as:

\begin{equation}
 \label{ce_cat}
   L_{CE}^{cat}(p_{ic}, \hat{p}_{ic})=\frac{1}{N}\sum^{N}_{i=1}\sum^{C}_{c=1}p_{ic} \cdot log{\hat{p}_{ic}}
 \end{equation}

\noindent where $p_{ic}$ and $\hat{p}_{ic}$ are the true and predicted probabilities for the $i^{th}$ instance to belong to the $c^{th}$ category, and $C$ is the total number of categories of the column. Similarly, textual columns follow the categorical cross-entropy loss $L_{CE}^{text}$ defined as:

\small
\begin{equation}
 \label{ce_text}
   L_{CE}^{text}(p_{itv}, \hat{p}_{itv})=\frac{1}{M}\sum^{N}_{i=1}\sum^{T_i}_{t=1}\sum^{V}_{v=1}p_{itv} \cdot log{\hat{p}_{itv}}
 \end{equation}
\normalsize

\noindent where $p_{itv}$ and $\hat{p}_{itv}$ are the true and predicted probabilities for the $t^{th}$ token of the $i^{th}$ instance to belong to the $v^{th}$ token in the vocabulary, $V$ is the size of the vocabulary, $T_i$ is the total number of tokens for the $i^{th}$ instance, and $M$ is the total number of tokens across all instances. To simultaneously impute for a numeric, a categorical, and a textual column, we minimize the total loss $L_{total}$ defined as:

\begin{equation}
 \label{total}
   L_{total} = L_{MSE} + L_{CE}^{cat} + L_{CE}^{text}
 \end{equation}

\noindent with the individual losses defined in Equations (3-5).

Though we visualize the framework in Figure \ref{mtl} to impute one column of each type and conduct experiments in the same way, this setup can be easily adjusted to impute any combination of numeric, categorical, and textual columns. Moreover, this enables the imputation of multiple columns at once and increases TTITA's usability. As we replace the missing values in the input columns with default values and do not require training data to be a complete table, we lower the requirements to utilize TTITA in custom applications.

\section{Experiments}
\subsection{Datasets}
We use two datasets from Amazon Reviews \citep{ni2019justifying} and one dataset on car reviews\footnote{https://huggingface.co/datasets/florentgbelidji/car-reviews}. The two Amazon datasets are the software and the gift cards categories, separately, with the metadata. The software category has 459,050 rows and the gift cards category has 147,190 rows. We impute the review summary with the review rating, verified status, reviewer identifier, item feature, and review text. Review rating is an integer on a scale of 1 to 5, verified status is binary, reviewer identifier is multi-categorical, and review text is unstructured text. Although the reviewer identifier has no missing data, the one-hot encoding representation is highly sparse, with 374,877 and 128,873 different reviewers. Since item feature in the raw data is represented in the form of a list of sequences, we concatenate each list of sequences into a single sequence. In multi-task learning, we additionally select to impute the Unix review time (numeric column) and the review's main category (categorical column). 

The car reviews dataset contains 36,984 rows and we impute the review title column with the rating (numeric), vehicle title (categorical), and review (text) columns. In multi-task learning, we additionally impute for the day of month of the review (numeric) and the reviewer name (categorical). The missing value percentages for each column and all datasets are listed in Table \ref{tab:dataset}. For pre-processing, numeric columns are standardized, and categorical columns are one-hot encoded.

\begin{table}[t!]
    \caption{Missing Value Percentages. We list the percentage of missing values for each column in every dataset. A pseudo-random 10\% of the datasets are used for testing, so the missing value percentages for the test sets follow similarly.}
    \label{tab:dataset}
    \begin{subtable}[h]{\linewidth}
    \centering
    \resizebox{\textwidth}{!}{
      \begin{tabular}{cccc}
        \toprule
    Rating & Vehicle Title & Review & Review Title \\
        \midrule
    0 & 0 & 0.73\% & 0.01\% \\
      \bottomrule
      \end{tabular}}
       \caption{Car dataset}
       \label{tab:carmissing}
    \end{subtable}
    \hfill
    \begin{subtable}[h]{\linewidth}
        \centering
        \resizebox{\textwidth}{!}{
        \begin{tabular}{cccccc}
        \toprule
    Rating & Verified & ID & Review & Feature & Summary \\
        \midrule
    0 & 0 & 0 & 0.11\% & 39.85\% & 0.03\% \\
      \bottomrule
      \end{tabular}}
        \caption{Gift cards dataset}
        \label{tab:giftcardmissing}
     \end{subtable}
     \hfill
     \begin{subtable}[h]{\linewidth}
        \centering
        \resizebox{\textwidth}{!}{
          \begin{tabular}{cccccc}
        \toprule
    Rating & Verified & ID & Review & Feature & Summary \\
        \midrule
    0 & 0 & 0 & 0.01\% & 10.29\% & 0.01\% \\
      \bottomrule
      \end{tabular}}
        \caption{Software dataset}
        \label{tab:softwaremissing}
     \end{subtable}
\end{table}

\subsection{Settings}
For each dataset, the standard training-validation-testing split is 81\%-9\%-10\%, and we select the set of model weights with the lowest validation loss across 10 epochs. We adopt the basic english tokenizer from PyTorch and cap the vocabulary size at 20,000 \citep{nation1997vocabulary} for efficiency despite that any size is feasible. We use the Adam optimizer \citep{kingma2017adam} with an initial learning rate of 0.0004 and a batch size of 128 on a single NVIDIA V100 GPU. The decoder layer is repeated 6 times and the number of heads in the attention blocks is 4 should it be a divisor of the size of the context vector, or lower otherwise. The first FC layer in the feed forward block has an output dimension of 1024, and the dropout layers have a dropout rate of 0.1.

\begin{table*}[t!]
    \caption{Evaluation Results. TTITA outperformed baseline models and the utilization of multi-task learning showed improvement.}
    \label{tab:main}
    \begin{subtable}[h]{\textwidth}
        \centering
        \resizebox{\textwidth}{!}{
          \begin{tabular}{cccccccccc}
            \toprule
             & Mode & KNN & LSTM & GRU & Decoder & Mistral & Llama2 & TTITA & TTITA-MTL \\
            \midrule
            METEOR & 0.0478 & 0.0318 & 0.0523 & 0.0507 & 0.0478 & \textbf{0.0904} & 0.0767 & 0.0590 & 0.0512 \\
            ROUGE & 0.0969 & 0.0535 & 0.1018 & 0.1012 & 0.0969 & 0.0734 & 0.0768 & \textbf{0.1129} & 0.1008 \\
            BLEU & 0.0579 & 0.0361 & 0.0627 & 0.0626 & 0.0579 & 0.0425 & 0.0606 & \textbf{0.0713} & 0.0617\\
          \bottomrule
        \end{tabular}}
        \caption{Car dataset results.}
        \label{tab:maincar}
     \end{subtable}
    \hfill
    \begin{subtable}[h]{\textwidth}
        \centering
        \resizebox{\textwidth}{!}{
          \begin{tabular}{cccccccccc}
            \toprule
             & Mode & KNN & LSTM & GRU & Decoder & Mistral & Llama2 & TTITA & TTITA-MTL \\
            \midrule
            METEOR & 0.3613 & 0.2778 & 0.4155 & 0.4119 & 0.3613 & 0.1138 & 0.0552 & 0.4216 & \textbf{0.4255}\\
            ROUGE & 0.3914 & 0.3142 & 0.4590 & 0.4639 & 0.3914 & 0.0423 & 0.0809 & \textbf{0.4769} & 0.4729 \\
            BLEU & 0.3913 & 0.3020 & 0.4399 & 0.4416 & 0.3913 & 0.0249 & 0.0573 & 0.4517 & \textbf{0.4531}\\
          \bottomrule
        \end{tabular}}
        \caption{Gift Cards dataset results.}
        \label{tab:maingift}
     \end{subtable}
     \hfill
        \begin{subtable}[h]{\textwidth}
    \centering
    \resizebox{\textwidth}{!}{
      \begin{tabular}{cccccccccc}
        \toprule
         & Mode & KNN & LSTM & GRU & Decoder & Mistral & Llama2 & TTITA & TTITA-MTL\\
        \midrule
        METEOR & 0.1237 & 0.1388 & 0.1923 & 0.1953 & 0.1237 & 0.1056 & 0.0308 & 0.2080 & \textbf{0.2125}\\
        ROUGE & 0.1328 & 0.1600 & 0.2257 & 0.2256 & 0.1328 & 0.0449 & 0.0551 & 0.2456 & \textbf{0.2474}\\
        BLEU & 0.1323 & 0.1544 & 0.2071 & 0.2084 & 0.1323 & 0.0281 & 0.0403 & 0.2217 & \textbf{0.2255} \\
      \bottomrule
    \end{tabular}}
       \caption{Software dataset results.}
       \label{tab:mainsoft}
    \end{subtable}
\end{table*}

\begin{table*}[t!]
    \caption{Model Performance for Imputing the Review Text. Instead of imputing the summary or title as in Table \ref{tab:main}, we impute the review text which has a much longer sequence length, and TTITA outperformed LSTM and GRU by a large margin.}
    \label{tab:abl}
    \centering
      \begin{tabular}{c|ccc|ccc|ccc}
        \toprule
         & \multicolumn{3}{c|}{Car Dataset} & \multicolumn{3}{c|}{Gift Cards Dataset} & \multicolumn{3}{c}{Software Dataset} \\
        \midrule
        & LSTM & GRU & TTITA & LSTM & GRU & TTITA & LSTM & GRU & TTITA \\
        \midrule
        METEOR & 0.0085 & 0.0155 & \textbf{0.1175} & 0.0894 & 0.0814 & \textbf{0.1231} & 0.0301 & 0.0377 & \textbf{0.1029}\\
        ROUGE & 0.0125 & 0.0289 & \textbf{0.1048} & 0.1433 & 0.1319 & \textbf{0.1781} & 0.0594 & 0.0675 & \textbf{0.1547} \\
        BLEU & 0.0057 & 0.0099 & \textbf{0.0933} & 0.0855 & 0.0725 & \textbf{0.1265} & 0.0155 & 0.0184 & \textbf{0.0924}\\
      \bottomrule
      \end{tabular}
\end{table*}

\subsection{Baseline Models}
\textbf{Mode}: We select the target sequence from the training set with the highest frequency to be the imputed text for all test instances. K Nearest Neighbor (\textbf{KNN}): We one-hot encode categorical columns, featurize textual columns with the hashing vectorizer, and keep numeric columns as is. The features are normalized and concatenated to represent each data sample. At inference for a test sample, we select the target sequence of the training instance that is closest to the test sample (K=1) in Euclidean distance. No training is required. \textbf{LSTM} \citep{hochreiter1997long}: The initial hidden state and cell state are initialized with the context vector and the token embedding layer has the same size. \textbf{GRU} \citep{chung2014empirical}: The initial hidden state is initialized with the context vector and the token embedding layer also has the same size. \textbf{Decoder}: We train a transformer decoder on the target column without having other columns as input, leaving out the context vector. This examines the significance of the context vector and verifies if the target textual column is self-sufficient for imputation. The decoder follows the same hyperparameter setup as TTITA. LLMs: We inquire the llama2-7b-chat model (\textbf{Llama2}) \citep{touvron2023llama} and the Mistral-7B-Instruct-v0.2 model (\textbf{Mistral}) \citep{jiang2023mistral} with the following prompt: \emph{Give me the [target column name] for a [dataset name] with the following values: A [column 1 name] of [column 1 value], a [column 2 name] of [column 2 value], ..., and a [last column name] of [last column value]. Please output the [target column name] and nothing else.} 

\subsection{Evaluation Metrics}
To evaluate and compare the performance of the models above, we utilize the following evaluation metrics on the ground truth $G$, and the imputed sequence $I$.

\textbf{METEOR} \citep{banerjee2005meteor}: We calculate the average METEOR score between all $G$ and $I$ pairs, which applies unigram matching and balances between precision and recall. Additionally, the metric takes into account how well the terms are ordered between the pair. \textbf{ROUGE} \citep{lin2004rouge}: We calculate the average unigram ROUGE F1 score between each $G$ and $I$ pair to quantify the level of overlapping between the target and imputed sequences. Precision captures the portion of correct tokens in $I$ while recall captures that in $G$. The F1-score as a whole marks how well the tokens overlap between the two sequences. \textbf{BLEU} \citep{papineni2002bleu}: We calculate the average unigram BLEU score between all $G$ and $I$ pairs to quantify the percentage of matching tokens but with the brevity penalty, which penalizes when $I$ is shorter than $G$.

\section{Results and Analysis}

\subsection{Performance}
We present the main evaluation results of all models in Table \ref{tab:main}. TTITA outperformed all baseline models on three datasets in every metric except the METEOR score on the car reviews dataset. With multi-task learning, TTITA-MTL achieved a slight improvement in METEOR (+0.39\%) and BLEU (+0.14\%) scores on the gift cards dataset and similar improvement in METEOR (+0.45\%), ROUGE (+0.18\%), and BLEU (+0.38\%) scores on the software dataset. A high score from mode imputation suggests that there is little variation among the target sequences in the gift cards dataset. Yet, TTITA is able to capture and impute the remaining diversity, demonstrating TTITA's superior capability than mode imputation and other competitive baseline models including the LSTM and GRU. With the two top-performing models, we conduct another experiment when the imputed column has a longer sequence length, and the results are shown in Table \ref{tab:abl}. We observe that TTITA's improvement is more evident (4$\sim$8\% compared to 1$\sim$2\%) when the imputed column has a longer sequence length, which corresponds to the advantages of the attention mechanism in the transformer. As another baseline model, a transformer decoder is trained on the target column without other input columns and TTITA's better performance validates our use of the context vector.

\begin{figure}[t]
  \centering
  \includegraphics[width=\linewidth]{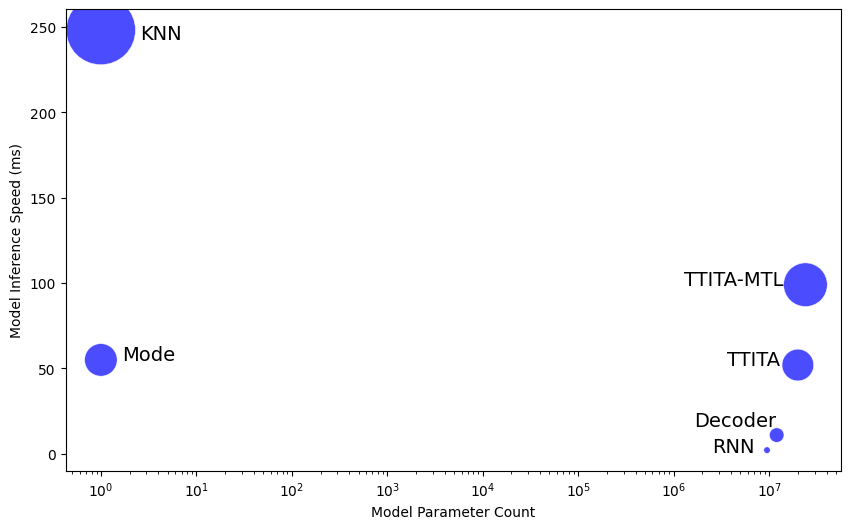}
  \caption{Model Parameter Count vs. Inference Speed. While being slower than LSTM and GRU, TTITA is comparable to Mode imputation and faster than KNN. Circle size is proportional to inference speed and RNN denotes LSTM and GRU.}
  \label{time}
\end{figure}

Figure \ref{time} visualizes the comparison between model parameter count and inference speed for all models, except Mistral and Llama2 having 7 billion parameters with 2$\sim$4 seconds of inference time. We observe that multi-task learning doubles the inference time while TTITA is comparable to mode imputation and faster than KNN. With quadratic time complexity in the attention mechanism and the additional context vector, TTITA performs slower than LSTM and GRU, and the decoder, respectively. Nonetheless, with an average inference time of 0.05 seconds, TTITA outperformed LSTM and GRU in evaluation metrics (Table \ref{tab:main}), and by a larger margin when the imputed column has a longer sequence length (Table \ref{tab:abl}). Additionally, we analyzed the importance of each column in Table \ref{tab:leaveoneout}. On the car reviews dataset, the review text has more importance than the rating which is equally important as the vehicle title.

\begin{figure*}
    \begin{subfigure}{0.32\linewidth}
         \centering
         \includegraphics[width=\linewidth]{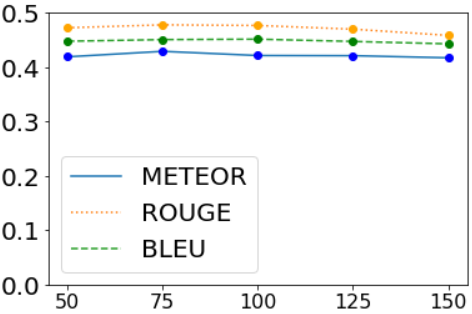}
         \caption{Numeric Column}
         \label{hyperparameter:numeric}
    \end{subfigure}
    \begin{subfigure}[b]{0.32\linewidth}
        \centering
         \includegraphics[width=\linewidth]{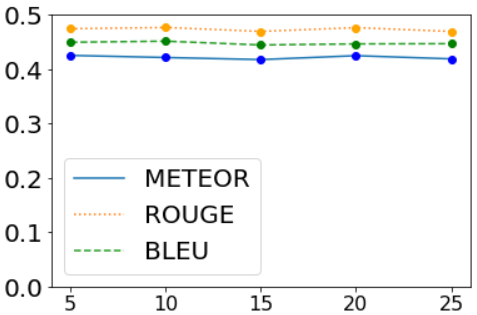}
         \caption{Categorical Column}
         \label{hyperparameter:categorical}
    \end{subfigure}
    \begin{subfigure}[b]{0.32\linewidth}
        \centering
         \includegraphics[width=\linewidth]{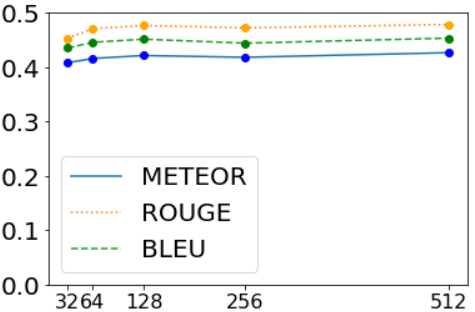}
         \caption{Textual Column}
         \label{hyperparameter:textual}
    \end{subfigure}
    \caption{Hyperparameter Analysis on Input Column Embedding Size. We investigate different embedding sizes for numeric ($d_n$), categorical ($d_c$), and textual ($d_t$) input columns. $d_n=100$, $d_c=10$, and $d_t=128$ resulted in better performance.}
    \label{hyperparameter}
\end{figure*}

\subsection{Comparison to LLMs}
Given the recent advances in language models and their ability in zero-shot learning, we investigate and observe that TTITA outperforms Mistral and Llama2-7b in most scenarios. Without being trained or fine-tuned on domain data, Llama2 generally suffers to impute the correct tokens. Some reviews have a long description of the product features in the Amazon datasets, and the lengthy prompt often confuses Llama2 into generating blanks. In some scenarios, the model attempts to summarize all the input columns instead of imputing the summary column. Nonetheless, when the dataset is not large enough for TTITA and multi-task learning to generalize well, LLMs can have a better performance, as we observe on the METEOR score of the car reviews dataset. 

Additionally, we conduct a similar analysis on ChatGPT. With the same prompt as Llama2 and Mistral, we observe similar results: the output length is much longer than the true summary, and the generated text also included other known attributes which are unnecessary in tabular data and require further post-processing in an application setting. This aligns with the observation \citep{li2023table} that language models are sub-optimal with tabular tasks. The repetitiveness also shows that ChatGPT is summarizing the input columns altogether, treating it as a summarization task, but our proposed method can impute columns of any relationship. In fact, we include a few other categorical and numerical input columns to show that this concept is scalable to numerous input columns of different types. Furthermore, the review text is encoded by an untrainable featurizer, which provides much less information on its summary as opposed to sequence-to-sequence prediction. This shows that we are not optimizing the performance by treating it as a summarization task, but rather using the dataset to demonstrate TTITA's performance. On the other hand, without being fine-tuned specifically, ChatGPT is unfamiliar with the summary text length in this dataset. Though length difference seems trivial, this can extend to other dataset-specific features that have meaningful implications in downstream tasks, such as the use of keywords suggesting a certain degree of familiarity of the subject. It is difficult for ChatGPT to impute the similar ``type'' of text even with prompt engineering. One can certainly provide further details on the dataset or examples for in-context learning \citep{dong2023survey}, but that involves a certain degree of human intervention/intelligence, and our focus is on developing a convenient framework for text imputation that requires no extra human effort. Additionally, there exists a limit on application programming interface calls which prevents systematic usage. Though TTITA does not contain open-world knowledge and requires training, it can impute unstructured textual columns of any relationship to the input columns and generate a ``similar'' text with an inference time of 0.05 seconds, making it a promising candidate for realistic applications.

\begin{table}[t!]
    \caption{Column Importance. On the car reviews dataset, we conduct a leave-one-out analysis on the input columns. Review text has slightly more importance than the rating and the vehicle title columns.}
    \label{tab:leaveoneout}
    \centering
    \resizebox{\linewidth}{!}{
      \begin{tabular}{c|cccc}
        \toprule
         & No Rating & No Vehicle Title & No Review \\
        \midrule
        METEOR & 0.0536 & 0.0566 & 0.0492\\
        ROUGE & 0.1056 & 0.1031 & 0.0996\\
        BLEU & 0.0642 & 0.0654 & 0.0601\\
      \bottomrule
      \end{tabular}}
\end{table}

\begin{table*}[t!]
    \caption{TTITA Prediction Examples on Amazon Reviews. Each row contains a sample from the test set and \emph{Prediction} has the imputed text. TTITA imputed tokens correctly and captured similar meanings. \emph{Other Columns} combines the other input columns for better presentation of this table (``,'' is used to separate each column attribute).}
    \resizebox{\textwidth}{!}{
  \begin{tabular}{|m{3.2cm}|m{11cm}|m{1.3cm}|m{1.4cm}|}
    \toprule
    Other Columns & Review Text & Summary & Prediction\\
    \midrule
    Software: 1, False, A3H5P8RBJKPCAC, NaN & I bought this Blu Ray thinking that it would be a good movie to watch in HD...Will Farrell comedy with dinosaurs...This has absolutely got to be the worst movie I have seen in the last 10 years or so. It lacked humor, a plot and anything remotely appealing. I was so disappointed that Will Farrell stouped this low to make a movie that was this bad. I am going to be taking it to Movie Stop and trading it in...that is if they will even buy it.& I'd give it a zero if I could & I am very disappointed...\\
    \hline
    Gift Card: 5, True, AR5ZTWU7JOQXL, NaN & This was a fast easy gift for a friend who has everything. She loved it! I will use these gift cards for birthdays in the future! & Great Gift! & great gift\\
    \bottomrule
  \end{tabular}}
\label{predictions}
\end{table*}

\subsection{Hyperparameter Analysis}
Using the gift cards dataset, we investigate in the change of model performance caused by different embedding sizes of the input columns in TTITA, as shown in Figure \ref{hyperparameter}. Within a selection of [50, 75, 100, 125, 150] neurons in the FC layer for numeric input columns, the size of 100 has the best performance. Between [5, 10, 15, 20, 25] for the embedding size of categorical input columns, the size of 10 performs the best. As for the output size of the hashing vectorizer, we choose from the values of [32, 64, 128, 256, 512] and a size of 128 has one of the better performances. These hyperparameter values can be easily adjusted for the data complexity of input columns, which makes TTITA suitable for custom applications.

\subsection{Qualitative Analysis}
This section dives into two text sequences imputed by TTITA, as shown in Table \ref{predictions}. Shorter columns, including the dataset the review was from, its rating, verified status, reviewer identifier, and item features, are combined into \emph{Other Columns} for better presentation of the table. The first example, from the software dataset, has low evaluation scores as the tokens do not overlap between the true and imputed sequences. However, the prediction summarizes the review well semantically. \emph{I'd give it a zero} suggests that it was below expectation when the lowest rating possible is one and that the reviewer is disappointed; the word $very$ matches the strong sentiment. The second example, from the gift cards dataset, has an accurately imputed summary of \emph{great gift}, except the exclamation mark. Through these examples, we observed the capability of TTITA to accurately impute textual columns both token-wise and meaning-wise. 

\section{Conclusion}
We introduced a novel approach (TTITA) to impute unstructured textual columns given other heterogeneous tabular columns using transformers. Supported by our experiment, the use of a context vector informatively benefited the decoder in end-to-end learning. TTITA achieved better performance than recurrent neural networks and by a larger margin when the imputed column has a longer sequence length. Extensive experimented also showed superior performance over large language models such as Mistral and Llama2 and common imputation methods such as KNN and mode imputation. Additionally, we leveraged multi-task learning to enable the simultaneous imputation of multiple heterogeneous columns and improved the performance on the target textual column. A comparison against ChatGPT on this task found TTITA to be more appropriate due to the advantage of training on domain data. Designed for minimal data pre-processing and human intervention, high generality and low inference times make TTITA a good candidate for realistic applications. Future work includes multi-domain adaptation to impute tabular datasets with a distribution shift, cross-lingual imputation involving multiple languages, bias removal on the imputed text, and exploration of fine-tuning LLMs for tabular data.

\bibliography{custom}
\bibliographystyle{acl_natbib}

\end{document}